\documentclass[11pt,a4paper]{article}
\usepackage[hyperref]{acl2019}
\usepackage{times}
\usepackage{latexsym}

\usepackage{url}

\usepackage{booktabs}
\usepackage{setspace}

\usepackage{amsmath,amssymb}
\usepackage{rotating}

\usepackage{multirow}

\usepackage{xcolor}
\usepackage{graphicx}

\aclfinalcopy 
\def\aclpaperid{1555} 

\title{Relational Word Embeddings}

\author{Jose Camacho-Collados \hspace{1cm} Luis Espinosa-Anke \hspace{1cm} Steven Schockaert \\
         School of Computer Science and Informatics \\
	    Cardiff University, United Kingdom\\
	     {\tt \{camachocolladosj,espinosa-ankel,schockaerts1\}@cardiff.ac.uk}}

\date{}

\begin{document}
\maketitle
\begin{abstract}
While word embeddings have been shown to implicitly encode various forms of attributional knowledge, the extent to which they capture relational information is far more limited. In previous work, this limitation has been addressed by incorporating relational knowledge from external knowledge bases when learning the word embedding. Such strategies may not be optimal, however, as they are limited by the coverage of available resources and conflate similarity with other forms of relatedness. As an alternative, in this paper we propose to encode relational knowledge in a separate word embedding, which is aimed to be complementary to a given standard word embedding. This relational word embedding is still learned from co-occurrence statistics, and can thus be used even when no external knowledge base is available. Our analysis shows that relational word vectors do indeed capture information that is complementary to what is encoded in standard word embeddings. 
\end{abstract}


\section{Introduction}

Word embeddings are paramount to the success of current natural language processing (NLP) methods. Apart from the fact that they provide a convenient mechanism for encoding textual information in neural network models, their importance mainly stems from the remarkable amount of linguistic and semantic information that they capture. For instance, the vector representation of the word \emph{Paris} implicitly encodes that this word is a noun, and more specifically a capital city, and that it describes a location in France. This information arises because word embeddings are learned from co-occurrence counts, and properties such as being a capital city are reflected in such statistics. However, the extent to which relational knowledge (e.g.\ \textit{Trump} was the successor of \textit{Obama}) can be learned in this way is limited.

Previous work has addressed this by incorporating external knowledge graphs \cite{DBLP:conf/cikm/XuBBGWLL14,enrichingAAAISS2015} or relations extracted from text \cite{chen2016webbrain}. 
However, the success of such approaches depends on the amount of available relational knowledge.
Moreover, they only consider well-defined discrete relation types (e.g.\ \textit{is the capital of}, or \textit{is a part of}), whereas the appeal of vector space representations largely comes from their ability to capture subtle aspects of meaning that go beyond what can be expressed symbolically. 
For instance, the relationship between \emph{popcorn} and  
\emph{cinema} is intuitively clear, but it is more subtle than the assertion that ``popcorn \emph{is located at} cinema'', 
which is how ConceptNet \cite{speer2017conceptnet}, for example, encodes this relationship\footnote{http://conceptnet.io/c/en/popcorn}. 

In fact, regardless of how a word embedding is learned, if its primary aim is to capture similarity, there are inherent limitations on the kinds of relations they can capture. 
For instance, such word embeddings can only encode similarity preserving relations (i.e.\ similar entities have to be related to similar entities) and it is often difficult to encode that $w$ is in a particular relationship while preventing the inference that words with similar vectors to $w$ are also in this relationship; e.g.\ \citeauthor{ziedcoling} \shortcite{ziedcoling} found that both (\emph{Berlin},\emph{Germany}) and (\emph{Moscow},\emph{Germany}) were predicted to be instances of the capital-of relation due to the similarity of the word vectors for \emph{Berlin} and \emph{Moscow}. Furthermore, while the ability to capture word analogies (e.g.\ \emph{king}-\emph{man}+\emph{woman}$\approx$\emph{queen}) emerged as a successful illustration of how word embeddings can encode some types of relational information \cite{mikolov2013linguistic}, the generalization of this interesting property has proven to be less successful than initially anticipated \cite{levy2014linguistic,linzen2016issues,rogers2017too}.

This suggests that relational information has to be encoded separately from standard similarity-centric word embeddings. One appealing strategy is to represent relational information by learning, for each pair of related words, a vector that encodes how the words are related. 
This strategy was first adopted by \citet{Turney:2005:MSS:1642293.1642475}, and has recently been revisited by a number of authors \cite{washio2018filling,DBLP:conf/acl/JameelSB18,DBLP:conf/coling/AnkeS18,DBLP:conf/emnlp/WashioK18,joshi2018pair2vec}. However, in many applications, word vectors are easier to deal with than vector representations of word pairs. 

The research question we consider in this paper is whether it is possible to learn word vectors that capture relational information. Our aim is for such \emph{relational word vectors} to be complementary to standard word vectors. To make relational information available to NLP models, it then suffices to use a standard architecture and replace normal word vectors by concatenations of standard and relational word vectors. In particular, we show that such relational word vectors can be learned directly from a given set of relation vectors. 



\section{Related Work}
\label{relatedWork}

\textbf{Relation Vectors.} A number of approaches have been proposed that are aimed at learning relation vectors for a given set of word pairs ($a$,$b$), based on sentences in which these word pairs co-occur. For instance, \citet{Turney:2005:MSS:1642293.1642475} introduced a method called Latent Relational Analysis (LRA), which relies on first identifying a set of sufficiently frequent lexical patterns and then constructs a matrix which encodes for each considered word pair ($a$,$b$) how frequently each pattern $P$ appears in between $a$ and $b$ in sentences that contain both words. Relation vectors are then obtained using singular value decomposition. More recently, \citet{DBLP:conf/acl/JameelSB18} proposed an approach inspired by the GloVe word embedding model \cite{glove2014} to learn relation vectors based on co-occurrence statistics between the target word pair $(a,b)$ and other words. Along similar lines, \newcite{DBLP:conf/coling/AnkeS18} learn relation vectors based on the distribution of words occurring in sentences that contain $a$ and $b$ by averaging the word vectors of these co-occurring words. Then, a conditional autoencoder is used to obtain lower-dimensional relation vectors. 

Taking a slightly different approach, \citet{washio2018filling} train a neural network to predict dependency paths from a given word pair. Their approach uses standard word vectors as input, hence relational information is encoded implicitly in the weights of the neural network, rather than as relation vectors (although the output of this neural network, for a given word pair, can still be seen as a relation vector). An advantage of this approach, compared to methods that explicitly construct relation vectors, is that evidence obtained for one word is essentially shared with similar words (i.e.\ words whose standard word vector is similar). Among others, this means that their approach can in principle model relational knowledge for word pairs that never co-occur in the same sentence.
A related approach, presented in \cite{DBLP:conf/emnlp/WashioK18}, uses lexical patterns, as in the LRA method, and trains a neural network to predict vector encodings of these patterns from two given word vectors. In this case, the word vectors are updated together with the neural network and an LSTM to encode the patterns. Finally, similar approach is taken by the Pair2Vec method proposed in \cite{joshi2018pair2vec}, where the focus is on learning relation vectors that can be used for cross-sentence attention mechanisms in tasks such as question answering and textual entailment. 

Despite the fact that such methods learn word vectors from which relation vectors can be predicted, it is unclear to what extent these word vectors themselves capture relational knowledge. In particular, the aforementioned methods have thus far only been evaluated in settings that rely on the predicted relation vectors. Since these predictions are made by relatively sophisticated neural network architectures, it is possible that most of the relational knowledge is still captured in the weights of these networks, rather than in the word vectors. Another problem with these existing approaches is that they are computationally very expensive to train; e.g.\ the Pair2Vec model is reported to need 7-10 days of training on unspecified hardware\footnote{\url{github.com/mandarjoshi90/pair2vec}}. In contrast, the approach we propose in this paper is computationally much simpler, while resulting in relational word vectors that encode relational information more accurately than those of the Pair2Vec model in lexical semantics tasks, as we will see in Section \ref{secExperiments}.

\smallskip\noindent\textbf{Knowledge-Enhanced Word Embeddings.}
Several authors have tried to improve word embeddings by incorporating external knowledge bases. For example, some authors have proposed models which combine the loss function of a word embedding model, to ensure that word vectors are predictive of their context words, with the loss function of a knowledge graph embedding model, to encourage the word vectors to additionally be predictive of a given set of relational facts \cite{DBLP:conf/cikm/XuBBGWLL14,enrichingAAAISS2015,chen2016webbrain}. 
Other authors have used knowledge bases in a more restricted way, by taking the fact that two words are linked to each other in a given knowledge graph as evidence that their word vectors should be similar \cite{DBLP:conf/naacl/FaruquiDJDHS15,speer2017conceptnet}. 
Finally, there has also been work that uses lexicons to learn word embeddings which are specialized towards certain types of lexical knowledge, such as hypernymy \cite{DBLP:conf/emnlp/NguyenKWV17,DBLP:conf/naacl/VulicM18},  antonymy \cite{liu2015learning,ono2015word} or a combination of various linguistic constraints \cite{mrkvsic2017semantic}.

Our method differs in two important ways from these existing approaches. First, rather than relying on an external knowledge base, or other forms of supervision, as in e.g.\ \cite{chen2016webbrain}, our method is completely unsupervised, as our only input consists of a text corpus. Second, whereas existing work has focused on methods for improving word embeddings, our aim is to learn vector representations that are \emph{complementary} to standard word embeddings. 

\section{Model Description}

We aim to learn representations that are complementary to standard word vectors and are specialized towards relational knowledge. To differentiate them from standard word vectors, they will be referred to as \textit{relational word vectors}. We write $\mathbf{e_w}$ for the relational word vector representation of $w$. 
The main idea of our method is to first learn, for each pair of closely related words $w$ and $v$, a relation vector $\mathbf{r_{wv}}$ that captures how these words are related, which we discuss in Section \ref{unsrelations}. In Section \ref{secRelationalWordVector} we then explain how we learn relational word vectors from these relation vectors. 

\subsection{Unsupervised Relation Vector Learning}\label{unsrelations}

Our goal here is to learn relation vectors for closely related words. For both the selection of the vocabulary and the method to learn relation vectors we mainly follow the initialization method of \newcite[\textsc{Relative}\textsubscript{init}]{relative2017ijcai} except for an important difference explained below regarding the symmetry of the relations. Other relation embedding methods could be used as well, e.g., \cite{DBLP:conf/acl/JameelSB18,DBLP:conf/emnlp/WashioK18,DBLP:conf/coling/AnkeS18,joshi2018pair2vec}, but this method has the advantage of being highly efficient.  
In the following we describe this procedure for learning relation vectors: we first explain how a set of potentially related word pairs is selected, and then focus on how relation vectors $\mathbf{r_{wv}}$ for these word pairs can be learned.   

\smallskip
\noindent \textbf{Selecting Related Word Pairs.}
Starting from a vocabulary $\mathcal{V}$ containing the words of interest (e.g.\ all sufficiently frequent words), as a first step we need to choose a set $\mathcal{R} \subseteq \mathcal{V} \times \mathcal{V}$ of potentially related words. For each of the word pairs in $\mathcal{R}$ we will then learn a relation vector, as explained below. To select this set $\mathcal{R}$, we only consider word pairs that co-occur in the same sentence in a given reference corpus. 
For all such word pairs, we then compute their strength of relatedness following \newcite{levy2015improving} by using a smoothed version of pointwise mutual information (PMI), where we use 0.5 as exponent factor. In particular, for each word $w \in \mathcal{V}$, the set $\mathcal{R}$ contains all sufficiently frequently co-occurring pairs $(w,v)$ for which $v$ is within the top-100 most closely related words to $w$, according to the following score:
\begin{equation}
\begin{aligned}
\textit{PMI}_{0.5}(u,v) &= \log \left(\frac{n_{wv} \cdot s_{**}}{n_{w*} \cdot s_{v*}}\right) 
\end{aligned}
\end{equation}
where $n_{wv}$ is the harmonically weighted\footnote{A co-occurrence in which there are $k$ words in between $w$ and $v$ then receives a weight of $\frac{1}{k+1}$.} number of times the words $w$ and $v$ occur in the same sentence within a distance of at most 10 words, and:
\begin{align*}
n_{w*} &= \sum_{u\in\mathcal{V}} n_{wu}; &
s_{v*} &= n_{v*}^{0.5}; &
s_{**} &= \sum_{u \in \mathcal{V}} s_{u*}
\end{align*}
This smoothed variant of PMI has the advantage of being less biased towards infrequent (and thus typically less informative) words. 

\smallskip
\noindent \textbf{Learning Relation Vectors.}
In this paper, we will rely on word vector averaging for learning relation vectors, which has the advantage of being much faster than other existing approaches, and thus allows us to consider a higher number of word pairs (or a larger corpus) within a fixed time budget. Word vector averaging has moreover proven surprisingly effective for learning relation vectors \cite{DBLP:conf/emnlp/WestonBYU13,DBLP:conf/conll/HashimotoSMT15,DBLP:conf/ranlp/FanCHG15,DBLP:conf/coling/AnkeS18}, as well as in related tasks such as sentence embedding \cite{wieting}. 

Specifically, to construct the relation vector $\mathbf{r_{wv}}$ capturing the relationship between the words $w$ and $v$ we proceed as follows. First, we compute a bag of words representation $\{(w_1,f_1),...,(w_n,f_n)\}$, where $f_i$ is the number of times the word $w_i$ occurs in between the words $w$ and $v$ in any given sentence in the corpus. The relation vector $\mathbf{r_{wv}}$ is then essentially computed as a weighted average:
\begin{equation}
\mathbf{r_{wv}} = \textit{norm}\left(\sum_{i=1}^n f_i \cdot \mathbf{w_i}\right)
\end{equation}
where we write $\mathbf{w_i}$ for the vector representation of $w_i$ in some given pre-trained word embedding, and $\textit{norm}(\mathbf{v})= \frac{\mathbf{v}}{\|\mathbf{v}\|}$.

In contrast to other approaches, we do not differentiate between sentences where $w$ occurs before $v$ and sentences where $v$ occurs before $w$. This means that our relation vectors are symmetric in the sense that $\mathbf{r_{wv}}=\mathbf{r_{vw}}$. This has the advantage of alleviating sparsity issues. While the directionality of many relations is important, the direction can often be recovered from other information we have about the words $w$ and $v$.  For instance, knowing that $w$ and $v$ are in a capital-of relationship, it is trivial to derive that ``$v$ is the capital of $w$'', rather than the other way around, if we also know that $w$ is a country.


\subsection{Learning Relational Word Vectors}\label{secRelationalWordVector}
\begin{figure}
\centering
\includegraphics[width=220pt,height=200pt]{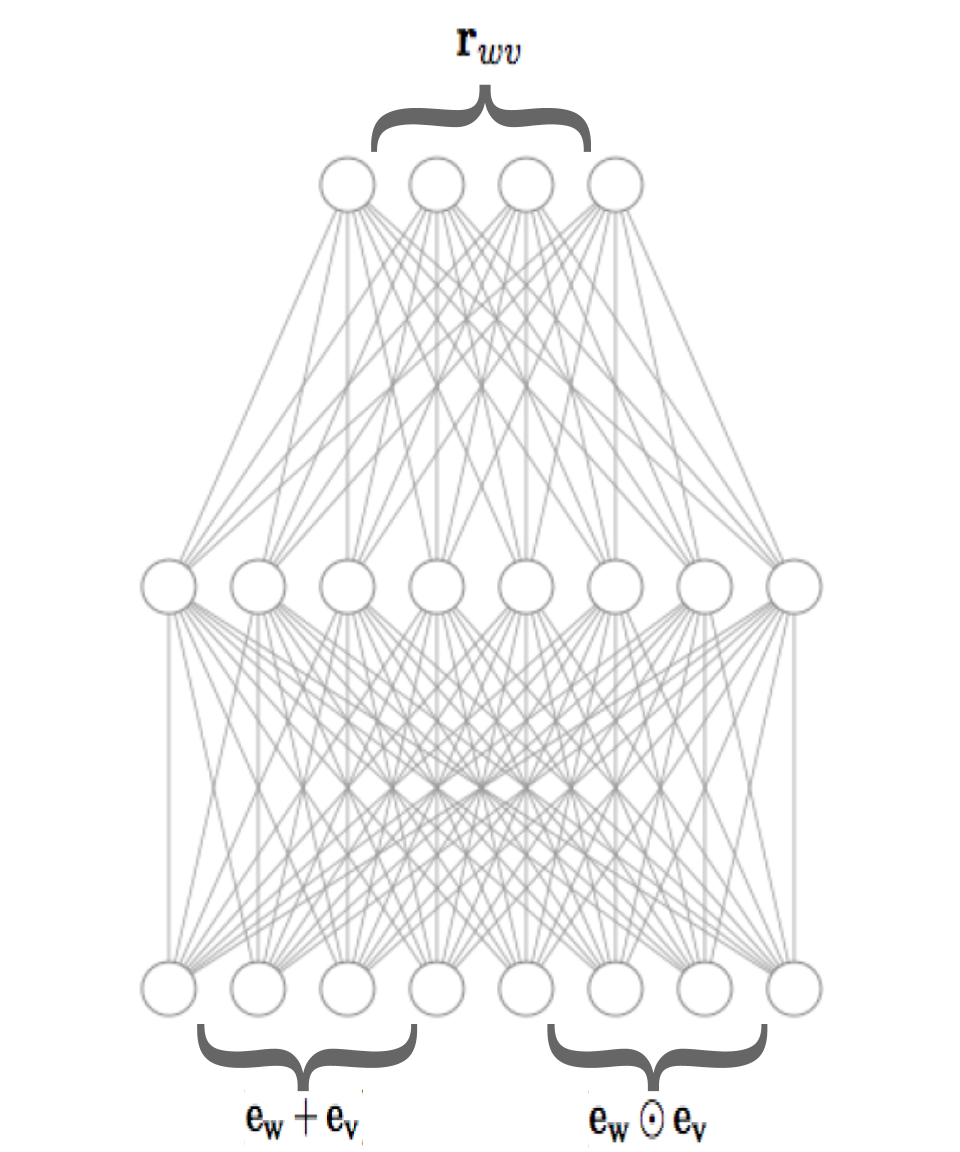}
\caption{
    Relational word embedding architecture. At the bottom of the figure, the input layer is constructed from the relational word embeddings $\mathbf{e}_{w}$ and $\mathbf{e}_{v}$, which are the vectors to be learnt. As shown at the top, we aim to predict the target relation vector $\mathbf{r}_{wv}$.
\label{fig:architecture}}
\end{figure}

The relation vectors $\mathbf{r_{wv}}$ capture relational information about the word pairs in $\mathcal{R}$. The relational word vectors will be induced from these relation vectors by encoding the requirement that $\mathbf{e_w}$ and $\mathbf{e_v}$ should be predictive of $\mathbf{r_{wv}}$, for each $(w,v)\in\mathcal{R}$. To this end, we use a simple neural network with one hidden layer,\footnote{More complex architectures could be used, e.g., \cite{joshi2018pair2vec}, but in this case we decided to use a simple architecture as the main aim of this paper is to encode all relational information into the word vectors, not in the network itself.} whose input is given by $(\mathbf{e_w} + \mathbf{e_v}) \oplus (\mathbf{e_w}\odot \mathbf{e_v})$, where we write $\oplus$ for vector concatenation and $\odot$ for the component-wise multiplication. Note that the input needs to be symmetric, given that our relation vectors are symmetric, which makes the vector addition and component-wise multiplication two straightforward encoding choices. Figure \ref{fig:architecture} depicts an overview of the architecture of our model. The network is defined as follows:
\begin{equation}
\label{eqrwe}
\begin{aligned}
\mathbf{i}_{wv} &= (\mathbf{e_w} + \mathbf{e_v}) \oplus (\mathbf{e_w}\odot \mathbf{e_v})\\
\mathbf{h}_{wv} &= f(\mathbf{X} \mathbf{i}_{wv} + \mathbf{a})\\
\mathbf{o}_{wv} &= f(\mathbf{Y} \mathbf{h}_{wv} + \mathbf{b})
\end{aligned}
\end{equation}
for some activation function $f$.
We train this network to predict the relation vector $\mathbf{r_{wv}}$, by minimizing the following loss:
\begin{equation}
\begin{aligned}
\mathcal{L} = \sum_{ (w,v) \in \mathcal{R}} &\Big(\mathbf{o}_{wv} - \mathbf{r}_{wv}\Big)^2 
\end{aligned}
\end{equation}
The relational word vectors $\mathbf{e_w}$ can be initialized using standard word embeddings trained on the same corpus.

\section{Experimental Setting}
\label{setting}

In what follows, we detail the resources and training details that we used to obtain the  relational word vectors. 

\smallskip
\noindent \textbf{Corpus and Word Embeddings.} We followed the setting of \newcite{joshi2018pair2vec} and used the English Wikipedia\footnote{Tokenized and lowercased dump of January 2018.} as input corpus. 
Multiwords (e.g. \textit{Manchester United}) were grouped together as a single token by following the same approach described in \newcite{DBLP:conf/nips/MikolovSCCD13}. As word embeddings, we used 300-dimensional FastText vectors \cite{FASTTEXT} trained on Wikipedia with standard hyperparameters. These embeddings are used as input to construct the relation vectors $\mathbf{r}_{wv}$ (see Section \ref{unsrelations}),\footnote{We based our implementation to learn relation vectors on the code available at \url{https://github.com/pedrada88/relative}} which are in turn used to learn relational word embeddings $\mathbf{e}_{w}$ (see Section \ref{secRelationalWordVector}). The FastText vectors are additionally used as our baseline word embedding model. 

\smallskip
\noindent \textbf{Word pair vocabulary.}
As our core vocabulary $\mathcal{V}$, we selected the $100,000$ most frequent words from Wikipedia. To construct the set of word pairs $\mathcal{R}$, for each word from $\mathcal{V}$, we selected the $100$ most closely related words (cf.\ Section \ref{unsrelations}), considering only consider word pairs that co-occur at least 25 times in the same sentence throughout the Wikipedia corpus. This process yielded relation vectors for 974,250 word pairs.

\smallskip
\noindent \textbf{Training.}
To learn our relational word embeddings we use the model described in Section \ref{secRelationalWordVector}. The embedding layer is initialized with the standard FastText 300-dimensional vectors trained on Wikipedia. 
The method was implemented in PyTorch, employing standard hyperparameters, using ReLU as the non-linear activation function $f$ (Equation \ref{eqrwe}). 
The hidden layer of the model was fixed to the same dimensionality as the embedding layer (i.e.\ 600). The stopping criterion was decided based on a small development set, by setting aside 1\% of the relation vectors. Code to reproduce our experiments, as well as pre-trained models and details of the implementation such as other network hyperparameters, are available at \url{https://github.com/pedrada88/rwe}. 

\section{Experimental Results}\label{secExperiments}

A natural way to assess the quality of word vectors is to test them in lexical semantics tasks. However, it should be noted that relational word vectors behave differently from standard word vectors, and  we should not expect the relational word vectors to be meaningful in unsupervised tasks such as semantic relatedness \cite{Turney10}. In particular, note that a high similarity between $\mathbf{e_w}$ and $\mathbf{e_v}$ should mean that relationships which hold for $w$ have a high probability of holding for $v$ as well. Words which are related, but not synonymous, may thus have very dissimilar relational word vectors.
Therefore, we test our proposed models on a number of different supervised tasks for which accurately capturing relational information is crucial to improve performance. 

\smallskip
\noindent \textbf{Comparison systems.} Standard FastText vectors, which were used to construct the relation vectors, are used as our main baseline. In addition, we also compare with the word embeddings that were learned by the Pair2Vec system\footnote{We used the pre-trained model of its official repository.} (see Section \ref{relatedWork}). We furthermore report the results of two methods which leverage knowledge bases to enrich FastText word embeddings: Retrofitting \cite{DBLP:conf/naacl/FaruquiDJDHS15} and Attract-Repel \cite{mrkvsic2017semantic}. Retrofitting 
exploits semantic relations from a knowledge base to re-arrange word vectors of related words such that they become closer to each other, whereas
Attract-Repel 
makes use of different linguistic constraints to move word vectors closer together or further apart depending on the constraint. For Retrofitting we make use of WordNet \cite{Fellbaum:98} as the input knowledge base, while for Attract-Repel we use the default configuration with all constraints from PPDB \cite{pavlick2015ppdb}, WordNet and BabelNet \cite{navigli2012babelnet}. All comparison systems are 300-dimensional and trained on the same Wikipedia corpus.

\begin{table*}[!t]
\footnotesize
\centering
\resizebox{\textwidth}{!}{
\renewcommand{\arraystretch}{1.15}
\begin{tabular}{llc|rrrr||rrrr}
   \multicolumn{1}{c}{\multirow{2}{*}{Encoding}} & \multicolumn{1}{c}{\multirow{2}{*}{Model}} & \multicolumn{1}{c}{\multirow{2}{*}{Reference}} & \multicolumn{4}{|c||}{\texttt{DiffVec}} & \multicolumn{4}{c}{\texttt{BLESS}}  \\ \cline{4-11} 
 & \multicolumn{1}{l|}{}  & \multicolumn{1}{l|}{} & \multicolumn{1}{c}{Acc.} & \multicolumn{1}{c}{F1} & \multicolumn{1}{c}{Prec.} & \multicolumn{1}{c||}{Rec.}  & \multicolumn{1}{c}{Acc.} & \multicolumn{1}{c}{F1} & \multicolumn{1}{c}{Prec.} & \multicolumn{1}{c}{Rec.} \\ \hline
 \multirow{5}{*}{Mult+Avg} 
  & \multicolumn{1}{l|}{\textsc{rwe}} & (This paper) &  \textbf{85.3}~~ & 	\textbf{64.2}~~ & 	65.1~~ & 	\textbf{64.5}~~ & \textbf{94.3} &	\textbf{92.8} &	\textbf{93.0} &	\textbf{92.6} \\
  & \multicolumn{1}{l|}{Pair2Vec} & \cite{joshi2018pair2vec} & 85.0~~ & 64.0~~ & 65.0~~ & 	\textbf{64.5}~~ & 91.2 & 	89.3 & 	88.9 & 	89.7 \\
    
  & \multicolumn{1}{l|}{FastText} & \cite{FASTTEXT}  & 84.2~~ & 	61.4~~ & 	62.6~~ & 	61.9~~ & 92.8 & 	90.4 & 	90.7 & 	90.2 \\
  \cline{2-11} 
 & \multicolumn{1}{l|}{Retrofitting$\dagger$} & \cite{DBLP:conf/naacl/FaruquiDJDHS15} & \textit{86.1}* & 	\textit{64.6}* & 	\textit{66.6}* & 	\textit{64.5}* &   90.6 & 	88.3 & 	88.1 & 	88.6  \\
  & \multicolumn{1}{l|}{Attract-Repel$\dagger$}  & \cite{mrkvsic2017semantic} & \textit{86.0}* & 	\textit{64.6}* & 	\textit{66.0}* & 	\textit{65.2}* &   91.2 & 	89.0 & 	88.8 & 	89.3 \\

  \hline
 \multirow{2}{*}{Mult+Conc}  
      & \multicolumn{1}{l|}{Pair2Vec}  & \cite{joshi2018pair2vec} & 84.8~~ & 	64.1~~ & 	\textbf{65.7}~~ & 	64.4~~ & 90.9 & 	88.8 & 	88.6 & 	89.1  \\
      & \multicolumn{1}{l|}{FastText}  & \cite{FASTTEXT} & 84.3~~ & 	61.3~~ & 	62.4~~ & 	61.8~~  & 92.9 &	90.6 &	90.8 &	90.4\\
      \hline
      \hline
 \multirow{1}{*}{Diff (only)}  
& \multicolumn{1}{l|}{FastText} & \cite{FASTTEXT} & 81.9~~ & 	57.3~~ & 	59.3~~ & 	57.8~~ & 88.5 & 	85.4 & 	85.7 & 	85.4   \\
      \hline

\end{tabular}
} 
\caption{Accuracy and macro-averaged F-Measure, precision and recall on BLESS and DiffVec. Models marked with $\dagger$ use external resources. The results with * indicate that WordNet was used for both the development of the model and the construction of the dataset. All models concatenate their encoded representations with the baseline vector difference of standard FastText word embeddings.}
\label{tab:blessdiffvec}
\end{table*}

\subsection{Relation Classification}
\label{relation_classification}

Given a pre-defined set of relation types and a pair of words, the relation classification task consists in selecting the relation type that best describes the relationship between the two words. As test sets we used DiffVec \cite{Vylomova2016} and BLESS\footnote{\url{http://clic.cimec.unitn.it/distsem}} \cite{baroni2011we}.
The DiffVec dataset includes 12,458 word pairs, covering fifteen relation types including hypernymy, cause-purpose or verb-noun derivations. On the other hand, BLESS includes semantic relations such as hypernymy, meronymy, and co-hyponymy.\footnote{Note that both datasets exhibit overlap in a number of relations as some instances from DiffVec were taken from BLESS.} BLESS includes a train-test partition, with 13,258 and 6,629 word pairs, respectively. 
This task is treated as a multi-class classification problem

As a baseline model (Diff), we consider the usual representation of word pairs in terms of their vector differences \cite{fu2014learning,roller2014inclusive,weeds2014learning}, using FastText word embeddings.  Since our goal is to show the complementarity of relational word embeddings with standard word vectors, for our method we concatenate the difference $\mathbf{w_j}-\mathbf{w_i}$ with the vectors $\mathbf{e_i}+\mathbf{e_j}$ and $\mathbf{e_i}\cdot \mathbf{e_j}$ (referred to as the Mult+Avg setting; our method is referred to as \textsc{rwe}). We use a similar representation for the other methods, simply replacing the relational word vectors by the corresponding vectors (but keeping the FastText vector difference). We also consider a variant in which the FastText vector difference is concatenated with $\mathbf{w_i}+\mathbf{w_j}$ and $\mathbf{w_i}\cdot \mathbf{w_j}$, which offers a more direct comparison with the other methods.
This goes in line with recent works that have shown how adding complementary features on top of the vector differences, e.g.\ multiplicative features \cite{vu2018integrating}, help improve the performance. 
Finally, for completeness, we also include variants where the average $\mathbf{e_i}+\mathbf{e_j}$ is replaced by the concatenation $\mathbf{e_i}\oplus\mathbf{e_j}$ (referred to as Mult+Conc), which is the encoding considered in \newcite{joshi2018pair2vec}. 

For these experiments we train a linear SVM classifier directly on the word pair encoding, 
performing a 10-fold cross-validation in the case of DiffVec, and using the train-test splits of BLESS. 




\paragraph{Results}  Table \ref{tab:blessdiffvec} shows the results of our relational word vectors, the standard FastText embeddings and other baselines on the two relation classification datasets (i.e.\ BLESS and DiffVec). Our model consistently outperforms the FastText embeddings baseline and comparison systems, with the only exception being the precision score for DiffVec. Despite being completely unsupervised, it is also surprising that our model manages to outperform the knowledge-enhanced embeddings of Retrofitting and Attract-Repel in the BLESS dataset. For DiffVec, let us recall that both these approaches have the unfair advantage of having had WordNet as source knowledge base,  used both to construct the test set and to enhance the word embeddings. 
In general, the improvement of \textsc{rwe} over standard word embeddings suggests that our vectors capture relations in a way that is compatible to standard word vectors (which will be further discussed in Section \ref{secNNsrelations}). 




\begin{table*}[t]
\renewcommand{\arraystretch}{1.15}
\setlength{\tabcolsep}{5pt}
\resizebox{\textwidth}{!}{
\small
{
\begin{tabular}{l|r|rrrrrrrr||r}
   \multicolumn{1}{c}{\multirow{2}{*}{Model}} & \multicolumn{9}{|c||}{\texttt{McRae Feature Norms}} & \multicolumn{1}{c}{\multirow{2}{*}{\texttt{QVEC}}}  \\ \cline{2-10} 
 & 
 \multicolumn{1}{c|}{Overall} & \multicolumn{1}{c}{metal} & \multicolumn{1}{c}{is\_small} & \multicolumn{1}{c}{is\_large}  & \multicolumn{1}{c}{animal} & \multicolumn{1}{c}{is\_edible} &  \multicolumn{1}{c}{wood}  & \multicolumn{1}{c}{is\_round} & \multicolumn{1}{c||}{is\_long}    \\ \hline
   \multicolumn{1}{l|}{\textsc{rwe}} & \textbf{55.2}  &\textbf{73.6}  & 46.7  & \textbf{45.9} & 89.2 & 61.5 & \textbf{38.5} & \textbf{39.0} & 46.8 & \textbf{55.4}~~ \\
   \multicolumn{1}{l|}{Pair2Vec} & 55.0  & 71.9  & \textbf{49.2}  & 43.3 & 88.9 & 68.3 & 37.7 & 35.0 & 45.5 & 52.7~~ \\

  \hline
 \multicolumn{1}{l|}{Retrofitting$\dagger$}  & 50.6  & 72.3  & 44.0  & 39.1 & \textbf{90.6} & \textbf{75.7} & 15.4 & 22.9 & 44.4 & \textit{56.8}* \\
  \multicolumn{1}{l|}{Attract-Repel$\dagger$}  & 50.4  & 73.2  & 44.4  & 33.3 & 88.9 & 71.8 & 31.1 & 24.2 & 35.9 & \textit{55.9}* \\
     \hline
      \hline   
  \multicolumn{1}{l|}{FastText} & 54.6  & 72.7  & 48.4  & 45.2 & 87.5 & 63.2 & 33.3 & \textbf{39.0} & \textbf{47.8} & 54.6~~ \\

\end{tabular}
}
}
\caption{Results on the McRae feature norms dataset (Macro F-Score) and QVEC (correlation score). Models marked with $\dagger$ use external resources. The results with * indicate that WordNet was used for both the development of the model and the construction of the dataset. 
}
\label{tab:features}
\end{table*}

\subsection{Lexical Feature Modelling}
\label{attributes}

Standard word embedding models tend to capture semantic similarity rather well \cite{baroni2014don,levy2015improving}. However, even though other kinds of lexical properties may also be encoded \cite{gupta2015distributional}, they are not explicitly modeled. 
Based on the hypothesis that relational word embeddings should allow  us to model such properties in a more consistent and transparent fashion, we select the well-known McRae Feature Norms benchmark \cite{mcrae2005semantic} as testbed. This dataset\footnote{Downloaded from \url{https://sites.google.com/site/kenmcraelab/norms-data}} is composed of 541 words (or concepts), each of them associated with one or more features. For example, `a \textit{bear} is an \textit{animal}', or `a \textit{bowl} is \textit{round}'. As for the specifics of our evaluation, given that some features are only associated with a few words, we follow the setting of \newcite{DBLP:conf/acl/RubinsteinLSR15} and consider the eight features with the largest number of associated words. We carry out this evaluation 
by treating the task as a multi-class classification problem, where the labels are the word features. 
As in the previous task, we use a linear SVM classifier and perform 3-fold cross-validation. For each input word, the word embedding of the corresponding feature is fed to the classifier concatenated with its baseline FastText embedding. 

Given that the McRae Feature Norms benchmark is focused on nouns, we complement this experiment with a specific evaluation on verbs. To this end, we use the verb set of QVEC\footnote{\url{https://github.com/ytsvetko/qvec}} \cite{tsvetkov2015evaluation}, a dataset specifically aimed at measuring the degree to which word vectors capture semantic properties which has shown to strongly correlate with performance in downstream tasks such as text categorization and sentiment analysis. QVEC was proposed as an intrinsic evaluation benchmark for estimating the quality of word vectors, and in particular whether (and how much) they predict lexical properties, such as words belonging to one of the fifteen verb supersenses contained in WordNet \cite{Miller1995}. 
As is customary in the literature, we compute Pearson correlation with respect to these predefined semantic properties, and measure how well a given set of word vectors is able to predict them, with higher being better. For this task we compare the 300-dimensional word embeddings of all models (without concatenating them with standard word embeddings), as the evaluation measure only assures a fair comparison for word embedding models of the same dimensionality. 

\paragraph{Results} Table \ref{tab:features} shows the results on the McRae Feature Norms dataset\footnote{Both \textit{metal} and \textit{wood} correspond to \textit{made of} relations.} and QVEC. 
In the case of the McRae Feature Norms dataset, our relational word embeddings achieve the best overall results, although there is some variation for the individual features. 
%
These results suggest that attributional information is encoded well in our relational word embeddings. 
Interestingly, our results also suggest that Retrofitting and Attract-Repel, which use pairs of related words during training, may be too na\"{i}ve to capture the complex relationships proposed in these benchmarks. In fact, they perform considerably lower than the baseline FastText model. On the other hand, Pair2Vec, which we recall is the most similar to our model, yields slightly better results than the FastText baseline, but still worse than our relational word embedding model. This is especially remarkable considering its much lower computational cost.

As far as the QVEC results are concerned, our method is only outperformed by Retrofitting and Attract-Repel. Nevertheless, the difference is minimal, which is surprising given that these methods leverage the same WordNet resource which is used for the evaluation. 

\section{Analysis}
\label{sec:qualitative}

To complement the evaluation of our relational word vectors on lexical semantics tasks, in this section we provide a qualitative analysis of their intrinsic properties.


\begin{table*}
\begin{center}
\renewcommand{\arraystretch}{1.3}

\resizebox{\textwidth}{!}{
\begin{tabular}{cc|cc|cc|cc}
\multicolumn{2}{c|}{\textsc{{\Large{sphere}}}} & \multicolumn{2}{c|}{\textsc{{\Large{philology}}}} & \multicolumn{2}{c|}{\textsc{{\Large{assassination}}}} &
\multicolumn{2}{c}{\textsc{{\Large{diversity}}}}
\\ \hline  \hline
\multicolumn{1}{c}{\textsc{rwe}} & \multicolumn{1}{c|}{FastText} & \multicolumn{1}{c}{\textsc{rwe}} & \multicolumn{1}{c|}{FastText} & \multicolumn{1}{c}{\textsc{rwe}} & \multicolumn{1}{c|}{FastText} & \multicolumn{1}{c}{\textsc{rwe}} & \multicolumn{1}{c}{FastText}\\ \hline
{\Large{rectangle}} & {\Large{spheres}} & {\Large{metaphysics}} & {\Large{philological}} & {\Large{riot}} & {\Large{assassinate}} & {\Large{connectedness}} & {\Large{cultural\_diversity}} \\
{\Large{conic}} & {\Large{spherical}} & {\Large{pedagogy}} & {\Large{philologist}} & {\Large{premeditate}} & {\Large{attempt}} & {\Large{openness}} & {\Large{diverse}}\\
{\Large{hexagon}} & {\Large{dimension}} & {\Large{docent}} & {\Large{literature}} & {\Large{bombing}} & {\Large{attempts}} & {\Large{creativity}} & {\Large{genetic\_diversity}}\\ \hline \hline
\multicolumn{2}{c|}{\textsc{{\Large{intersect}}}} & \multicolumn{2}{c|}{\textsc{{\Large{behaviour}}}} & \multicolumn{2}{c|}{\textsc{{\Large{capability}}}} &
\multicolumn{2}{c}{\textsc{{\Large{execute}}}}
\\ \hline  \hline
\multicolumn{1}{c}{\textsc{rwe}} & \multicolumn{1}{c|}{FastText} & \multicolumn{1}{c}{\textsc{rwe}} & \multicolumn{1}{c|}{FastText} & \multicolumn{1}{c}{\textsc{rwe}} & \multicolumn{1}{c|}{FastText} & \multicolumn{1}{c}{\textsc{rwe}} & \multicolumn{1}{c}{FastText}\\ \hline
{\Large{tracks}} & {\Large{intersection}} & {\Large{aggressive}} & {\Large{behaviour}} & {\Large{refueling}} & {\Large{capabilities}} & {\Large{murder}} & {\Large{execution}} \\
{\Large{northbound}} & {\Large{bisect}} & {\Large{detrimental}} & {\Large{behavioural}} & {\Large{miniaturize}} & {\Large{capable}} & {\Large{interrogation}} & {\Large{executed}} \\
{\Large{northwesterly}} & {\Large{intersectional}} & {\Large{distasteful}} & {\Large{misbehaviour}} & {\Large{positioning}} & {\Large{survivability}} & {\Large{incarcerate}} & {\Large{summarily\_executed}} \\
\end{tabular}
}
\end{center}
\caption{\label{tab:closestwords} Nearest neighbours for selected words in our relational word embeddings (\textsc{rwe}) and FastText embeddings 
}
\end{table*}

\subsection{Word Embeddings: Nearest Neighbours}
\label{secNNswords}

First, we provide an analysis based on the nearest neighbours of selected words in the vector space. Table \ref{tab:closestpairs} shows nearest neighbours of our relational word vectors and the standard FastText embeddings.\footnote{Recall from Section \ref{setting} that both were trained on Wikipedia with the same dimensionality, i.e., 300.} The table shows that our model captures some subtle properties, which are not normally encoded in knowledge bases. For example, geometric shapes are clustered together around the \textit{sphere} vector, unlike in FastText, where more loosely related  words such as ``dimension'' are found. This trend can easily be observed as well in the \textit{philology} and \textit{assassination} cases. 

In the bottom row, we show cases where relational information is somewhat confused with collocationality, leading to undesired clusters, such as \textit{intersect} being close in the space with ``tracks'', or \textit{behaviour} with ``aggressive'' or ``detrimental''. These examples thus point towards a clear direction for future work, in terms of explicitly differentiating collocations from other relationships.

\subsection{Word Relation Encoding}
\label{secNNsrelations}

Unsupervised learning of analogies has proven to be one of the strongest selling points of word embedding research. Simple vector arithmetic, or pairwise similarities \cite{levy2014linguistic}, can be used to capture a surprisingly high number of semantic and syntactic relations. We are thus interested in exploring \textit{semantic clusters} 
as they emerge when encoding relations using our relational word vectors. Recall from Section \ref{secRelationalWordVector} that relations are encoded using addition and point-wise multiplication of word vectors. 


Table \ref{tab:closestpairs} shows, for a small number of selected word pairs, the top nearest neighbors that were unique to our 300-dimensional relational word vectors. Specifically, these pairs were not found among the top 50 nearest neighbors for the FastText word vectors of the same dimensionality, using the standard vector difference encoding. Similarly, we also show the top nearest neighbors that were unique to the FastText word vector difference encoding.
As can be observed, our relational word embeddings can capture interesting relationships which go beyond what is purely captured by similarity. For instance, for the pair ``innocent-naive'' our model includes similar relations such as \textit{vain-selfish}, \textit{honest-hearted} or \textit{cruel-selfish} as nearest neighbours, compared with the nearest neighbours of standard FastText embeddings which are harder to interpret.

\begin{table*}
\begin{center}
{
\resizebox{\textwidth}{!}{
\begin{tabular}{cc|cc|cc}
\multicolumn{2}{c|}{\textsc{\large{innocent-naive}}} & \multicolumn{2}{c|}{\textsc{\large{poles-swedes}}} & \multicolumn{2}{c}{\textsc{\large{shoulder-ankle}}} \\ \hline  \hline
\multicolumn{1}{c}{\textsc{rwe}} & \multicolumn{1}{c|}{FastText} & \multicolumn{1}{c}{\textsc{rwe}} & \multicolumn{1}{c|}{FastText} & \multicolumn{1}{c}{\textsc{rwe}} & \multicolumn{1}{c}{FastText} \\ \hline
vain-selfish & murder-young  & lithuanians-germans & polish-swedish & wrist-knee & oblique-ligament \\
honest-hearted & imprisonment-term  & germans-lithuanians  & poland-sweden & thigh-knee & pick-ankle\_injury \\
cruel-selfish & conspiracy-minded  & russians-lithuanians  & czechoslovakia-sweden & neck-knee & suffer-ankle\_injury \\
\hline \hline
\multicolumn{2}{c|}{\textsc{\large{shock-grief}}} & \multicolumn{2}{c|}{\textsc{\large{strengthen-tropical\_cyclone}}} & \multicolumn{2}{c}{\textsc{\large{oct-feb}}} \\ \hline  \hline
\multicolumn{1}{c}{\textsc{rwe}} & \multicolumn{1}{c|}{FastText} & \multicolumn{1}{c}{\textsc{rwe}} & \multicolumn{1}{c|}{FastText} & \multicolumn{1}{c}{\textsc{rwe}} & \multicolumn{1}{c}{FastText} \\ \hline
anger-despair & overcome-sorrow & intensify-tropical\_cyclone & name-tropical\_cyclones  & aug-nov & doppler-wheels\\
anger-sorrow & overcome-despair & weaken-tropical\_storm & bias-tropical\_cyclones & sep-nov & scanner-read \\
anger-sadness & moment-sadness & intensify-tropical\_storm & scheme-tropical\_cyclones & nov-sep & ultrasound-baby 
\end{tabular}
}

}
\end{center}
\caption{\label{tab:closestpairs} Three nearest neighbours for selected word pairs using our relational word vector's relation encoding (\textsc{rwe}) and the standard vector difference encoding of FastText word embeddings. In each column only the word pairs which were on the top 50 NNs of the given model but not in the other are listed. Relations which include one word from the original pair were not considered.
}
\end{table*}

Interestingly, even though not explicitly encoded in our model, the table shows some examples that highlight one property that arises often, which is the ability of our model to capture co-hyponyms as relations, e.g., \textit{wrist-knee} and \textit{anger-despair} as nearest neighbours of ``shoulder-ankle'' and ``shock-grief'', respectively. Finally, one last advantage that we highlight is the fact that our model seems to perform implicit disambiguation by balancing a word's meaning with its paired word. For example, the ``oct-feb'' relation vector correctly brings together other \textit{month abbreviations} in our space, whereas in the FastText model, its closest neighbour is `doppler-wheels', a relation which is clearly related to another sense of \textit{oct}, namely its use as an acronym to refer to `optical coherence tomography' (a type of x-ray procedure that uses the \textit{doppler} effect principle).


\subsection{Lexical Memorization}
\label{lexicalmemorization}

One of the main problems of word embedding models performing lexical inference (e.g.\ hypernymy) is lexical memorization. \newcite{Levyetal2015} found that the high performance of supervised distributional models in hypernymy detection tasks was due to a memorization in the training set of what they refer to as \textit{prototypical hypernyms}. These prototypical hypernyms are general categories which are likely to be hypernyms (as occurring frequently in the training set) regardless of the hyponym. For instance, these models could equally predict the pairs \textit{dog-animal} and \textit{screen-animal} as hyponym-hypernym pairs. To measure the extent to which our model is prone to this problem we perform a controlled experiment on the lexical split of the HyperLex dataset \cite{vulic2017hyperlex}. This lexical split does not contain any word overlap between training and test, and therefore constitutes a reliable setting to measure the generalization capability of embedding models in a controlled setting \cite{shwartz2016improving}. In HyperLex, each pair is provided by a score which measures the strength of the hypernymy relation.

For these experiments we considered the same experimental setting as described in Section \ref{setting}. In this case we only considered the portion of the HyperLex training and test sets covered in our vocabulary\footnote{Recall from Section \ref{setting} that this vocabulary is shared by all comparison systems.} and used an SVM regression model over the word-based encoded representations. Table \ref{tab:hyperlex} shows the results for this experiment. Even though the results are low overall (noting e.g.\ that results for the random split are in some cases above 50\% as reported in the literature), our model can clearly generalize better than other models. Interestingly, methods such as Retrofitting and Attract-Repel perform worse than the FastText vectors. This can be attributed to the fact that these models have been mainly tuned towards similarity, which is a feature that loses relevance in this setting. Likewise, the relation-based embeddings of Pair2Vec do not help, probably due to the high-capacity of their model, which makes the word embeddings less informative. 

\begin{table}[!t]
\small
\centering
\renewcommand{\arraystretch}{1.15}
\begin{tabular}{ll|rr}
   \multicolumn{1}{c}{Encoding} & \multicolumn{1}{c}{Model}  & \multicolumn{1}{c}{$r$} & \multicolumn{1}{c}{$\rho$} \\ \hline
 \multirow{5}{*}{Mult+Avg} 
  & \multicolumn{1}{l|}{\textsc{rwe}} &  \textbf{38.8}~~ & \textbf{38.4}~~  \\
  & \multicolumn{1}{l|}{Pair2Vec} &  28.3~~ & 26.5~~  \\
    
  & \multicolumn{1}{l|}{FastText} &  37.2~~ & 35.8~~ \\
  \cline{2-4} 
 & \multicolumn{1}{l|}{Retrofitting$\dagger$} &   \textit{29.5*} & \textit{28.9*}  \\
  & \multicolumn{1}{l|}{Attract-Repel$\dagger$}   &  \textit{29.7*} & \textit{28.9*}  \\

  \hline
 \multirow{2}{*}{Mult+Conc}  
      & \multicolumn{1}{l|}{Pair2Vec}   &  29.8~~ & 30.0~~  \\
      & \multicolumn{1}{l|}{FastText} &  35.7~~ & 33.3~~ \\
      \hline
      \hline
 \multirow{1}{*}{Diff (only)}  
& \multicolumn{1}{l|}{FastText} &  29.9~~ & 30.1~~  \\
      \hline

\end{tabular}
\caption{Pearson ($r$) and Spearman ($\rho$) correlation on a subset of the HyperLex lexical split. Models marked with $\dagger$ use external resources. All models concatenate their encoded representations with the baseline vector difference of standard FastText word embeddings.}
\label{tab:hyperlex}
\end{table}

\section{Conclusions}


We have introduced the notion of relational word vectors, and presented an unsupervised method for learning such representations. Parting ways from previous approaches where relational information was either encoded in terms of relation vectors (which are highly expressive but can be more difficult to use in applications), represented by transforming standard word vectors (which capture relational information only in a limited way), or by taking advantage of external knowledge repositories, 
we proposed to learn an unsupervised word embedding model that is tailored specifically towards modelling relations. Our model is intended to capture knowledge which is complementary to that of standard similarity-centric embeddings, and can thus be used in combination. 

We tested the complementarity of our relational word vectors with standard FastText word embeddings on several lexical semantic tasks, capturing different levels of relational knowledge. The evaluation indicates that our proposed method indeed results in representations that capture relational knowledge in a more nuanced way. 
For future work, we would be interested in further exploring the behavior of neural architectures for NLP tasks which intuitively would benefit from having access to relational information, e.g., text classification \cite{DBLP:conf/coling/AnkeS18, relative2017ijcai} and other language understanding tasks such as natural language inference or reading comprehension, in the line of \newcite{joshi2018pair2vec}. 



\smallskip
\noindent\textbf{Acknowledgments.} 
Jose Camacho-Collados and Steven Schockaert were supported by ERC Starting Grant 637277.

\bibliography{commonsense,wordembedding,entity}
\bibliographystyle{acl_natbib}

\end{document}